\documentclass[12pt,a4paper]{article}
\usepackage{amsmath,amsfonts}
\usepackage{algorithmic}
\usepackage{algorithm}
\usepackage{array}
\usepackage[caption=false,font=small,labelfont=bf]{subfig}
\usepackage{textcomp}
\usepackage{stfloats}
\usepackage{url}
\usepackage{verbatim}
\usepackage{graphicx}
\usepackage{cite}
\usepackage{hyperref}
\usepackage[margin=2.5cm]{geometry}
\usepackage{tikz}
\usepackage{booktabs}
\usepackage{tabularx}

\begin{document}

\title{DABACO: A Multi-Camera Dataset and Benchmark for Screen Localization and Pointing Estimation}

\author{
    Óscar Gómez-Cárdenes, José Gil Marichal-Hernández, \\ Juan Manuel Martín-Doñas \\
    \textit{Universidad de La Laguna (ULL)} \\
    \thanks{This work was supported by the Universidad de La Laguna (ULL) Plan Propio de Investigación 2025.}
}

\maketitle

\begin{abstract}

Screen localization and pointing estimation are key components in the development of low-cost interactive devices. Yet developing and evaluating these algorithms requires realistic data: synthetic captures cannot fully reproduce the optical distortion, rolling shutter, motion blur, and display processing of a physical acquisition, and most existing datasets provide isolated static images rather than the video required to assess continuous pointing. In this paper, we introduce the DABACO Dataset. Developed within the DABACO (Dispositivo Apuntador de BAjo COste, or Low-Cost Pointing Device) project, this dataset supports the development and evaluation of screen detection and camera-based pointing algorithms for embedded systems. The dataset comprises video sequences captured with multiple low-cost camera sensors, including monochrome global-shutter and color rolling-shutter modules, operated on representative embedded platforms such as the Raspberry Pi 4B and ESP32-S3.

We present a systematic annotation pipeline combining temporary visual watermarking, optical-flow tracking, manual verification, and marker removal. Both the original marked captures and the marker-free images, together with explicit corner annotations and modification masks, are released to support auditing and the study of potential reconstruction bias.

In addition, we release an open-source evaluation toolkit with two reference baselines: a classical screen-detection pipeline based on edge and contour geometry, and a closed-vocabulary, off-the-shelf YOLOv8 segmentation model evaluated without dataset-specific training. Both baselines are evaluated on a general-purpose computer using Intersection over Union, corner error, and pointing error, establishing initial reference results for future implementations optimized for embedded edge platforms. Overall, DABACO addresses a gap in existing resources and is intended to support the development and evaluation of new low-cost screen-localization and pointing systems.

\end{abstract}


\section{Introduction}
\label{sec:intro}

Interactive pointing devices, including lightguns, have been widely used with classic arcade machines and home consoles. Light-based pointing long predates the video game industry: commercial light-gun games appeared as early as the 1930s, and by the 1970s the principle had moved into the living room, with the television receiver itself acting as the target~\cite{griffith1934lightgun,baer1972television}. In that lineage, classic lightguns exploited the raster-scanning behavior of Cathode-Ray Tube (CRT) displays to infer the pointed location from the timing of photodiode responses: the device reports no position of its own, it only times the light it receives against the raster~\cite{nesdevzapper}. This operating principle is not directly applicable to modern flat-panel displays, whose image formation, buffering, scaling, and display-processing pipelines differ substantially from CRT raster scanning.

Camera-based solutions have since enabled pointing on modern displays. The Sinden Lightgun uses an integrated camera to detect a high-contrast border rendered around the display content~\cite{sinden2022control,sinden2024website}; although effective with flat-panel displays, this approach modifies the displayed image and relies on a wired connection and host-side processing. Other pointing systems use different references: air mice primarily rely on inertial sensors to estimate relative cursor motion, while the Nintendo Wii Remote uses an infrared camera to observe an external Sensor Bar placed near the display~\cite{lee2008wiimote}. These systems illustrate the trade-off between relative motion sensing, external references, and visual localization of the display itself, and they motivate the need for low-cost pointing systems that require neither a modified image nor additional hardware in the environment.

Estimating the geometry and pointing coordinates of an arbitrary display from a low-cost camera---without a visible border, an infrared Sensor Bar, or other external references---is the problem addressed by the DABACO (\textit{Dispositivo Apuntador de BAjo COste}, Spanish for Low-Cost Pointing Device) project. Accurate geometric extraction is critical for this task, since small corner deviations severely distort the perspective projection required for precise pointing. Moreover, whereas most existing datasets provide isolated static images for object detection, the temporal continuity of video is essential to evaluate the tracking stability of continuous pointing, and the strict resource constraints of edge devices heavily influence both acquisition and inference. Developing and evaluating such algorithms therefore requires standardized data captured natively on heterogeneous, low-cost embedded hardware. The closest existing resources, however, fall short: planar-object tracking and video-homography benchmarks provide video with four-corner annotations but target planar objects in general---a category that may include screens---without annotating the active area of a display or evaluating pointing; screen-recapture and screen-presence datasets do involve real displays but annotate image-level classes or bounding boxes rather than the active area; and large pointing datasets annotate human pointing in a third-person view rather than the pointing coordinate of a camera-based device. Section~\ref{sec:related} details each of these families and Table~\ref{tab:related-datasets} summarizes the comparison.

To address this gap, this paper introduces the DABACO Dataset. The initial release provides controlled captures covering multiple display devices, visual contents, camera sensors, and acquisition settings, acquired with monochrome global-shutter and color rolling-shutter modules on representative embedded platforms such as the Raspberry Pi 4B and ESP32-S3. The main contributions of this work are threefold:
\begin{enumerate}
    \item We introduce the DABACO Dataset, a multi-camera video resource for screen localization and camera-based pointing estimation under heterogeneous acquisition conditions.
    \item We describe a systematic, semi-automated annotation procedure combining temporary ArUco markers, optical-flow tracking, and manual verification, followed by marker removal. Beyond this dataset, the procedure is intended to be reusable by researchers who need to produce comparable screen-corner annotations. The release includes geometric annotations, marker-based captures, marker-free inpainted images, and explicit modification masks to support auditing and the analysis of reconstruction artifacts.
    \item We provide an open-source evaluation toolkit with standardized metrics and two reference baselines: a classical computer-vision pipeline and a closed-vocabulary, off-the-shelf YOLOv8 segmentation baseline. Results obtained on a general-purpose computer establish initial reference values for future implementations optimized for embedded edge platforms.
\end{enumerate}

The dataset and the evaluation toolkit, together with the per-frame results of both baselines, are released in open access.\footnote{Dataset: \cite{dabaco_dataset_2026}. Source code and evaluation toolkit: \url{https://domondo.github.io/dabaco-dataset-toolkit/}.}

\section{Related Work}
\label{sec:related}

\subsection{Pointing Systems}

Interactive pointing systems rely on various reference mechanisms to estimate the user's target. Classic CRT lightguns exploited the raster-scanning behavior of the display to infer the pointed location from the timing of photodiode responses---the device reports no position of its own, it only times the light it receives against the raster---an approach incompatible with modern flat-panel displays. Consequently, alternative references have been adopted. Air mice and presentation controllers primarily use inertial sensors to provide relative cursor control rather than absolute coordinates. Infrared-reference systems, such as the Nintendo Wii Remote~\cite{lee2008wiimote}, estimate device pose by observing an external Sensor Bar placed near the display. While effective, this requires additional hardware in the environment. Camera-based lightguns, such as the Sinden Lightgun~\cite{sinden2022control}, eliminate the need for external hardware by detecting a high-contrast visual border explicitly rendered around the active display content, but this approach modifies the displayed image and typically relies on a wired connection and host-side processing.

It is important to distinguish between relative cursor control, general device pose estimation, and absolute screen localization. DABACO specifically targets absolute pointing coordinates obtained directly from the visual localization of the display. It aims to estimate the geometry and pointing coordinates of an arbitrary screen using a low-cost camera and embedded processing, without requiring a visible border, an infrared Sensor Bar, or other external physical references. The target architecture is intended to operate without permanent external references; the temporary markers used in this dataset are restricted to the annotation-generation process and are removed or explicitly masked in the released image variants.

\subsection{Display Localization and Screen--Camera Systems}

Beyond pointing devices, display geometry has also been considered in screen-localization, screen--camera communication, and robotic-interaction systems. The detection of display geometry from a camera has been claimed and studied outside the gaming literature: one patent protects the real-time estimation of a screen's coordinates and shape from a vision sensor~\cite{patent2017screendetect}, and another detects the screen area to guide a robotic arm~\cite{patent2020roboticarm}. In the academic literature, Snapscreen addresses the retrieval of a TV-stream frame from a projectively distorted and noisy camera query~\cite{skoryukina2017snapscreen}, and screen tracking has been used to enable the clinical translation of live ultrasound image analysis~\cite{screentracking2020}. Screen--camera communication systems such as PixNet~\cite{perli2010pixnet}, LightSync~\cite{hu2013lightsync}, and DisCo~\cite{jo2016disco} demonstrate the feasibility of extracting information from flat-panel displays with low-cost cameras, including under frame-rate mismatch and rolling-shutter acquisition. Their objective, however, is communication rather than display localization or pointing, so they are complementary rather than directly comparable to DABACO.

\subsection{Datasets for Homography and Screen Detection}

Several publicly available datasets and benchmarks address sub-problems related to screen localization, such as homography estimation and object detection. HPatches~\cite{balntas2017hpatches} provides image sequences with viewpoint and illumination changes and associated homography information; however, it is primarily designed for evaluating handcrafted and learned local features rather than physical-display localization or pointing. Similarly, the Homography Estimation Benchmark (HEB) and the associated Pi3D dataset~\cite{barath2023large} offer large-scale benchmarks for generic planar homography estimation across diverse scenes and viewpoints, but they are not specifically designed for screen localization, pointing-coordinate estimation, or low-cost embedded camera systems.

In the domain of object detection, resources such as the Screen Detection YOLO dataset (hosted on Mendeley Data)~\cite{su2023screen} and the Screen Detection v2 dataset (available via Roboflow Universe)~\cite{screen-detection-v2_dataset} provide annotated images of physical digital displays (e.g., laptops, monitors, phones) in various environments. For the two screen-detection datasets, ``physical displays'' refers to the presence of photographed display devices in the published samples; this attribute was verified by inspecting the released images. However, these datasets are tailored for bounding-box object detection on independent static images rather than video sequences. Furthermore, they often annotate the entire physical device structure (including bezels and stands) rather than the precise inner geometric boundaries of the active screen area. Consequently, they lack the temporal continuity, exact quadrilateral screen corners, and homography labels that are essential for accurate pointing estimation and evaluation.

Recent benchmarks close part of that distance and must be acknowledged explicitly. PlanarTrack~\cite{planartrack2023,planartrack2025} and the earlier POT-210 benchmark~\cite{pot2018} provide videos of planar objects with four-corner annotations, perspective variation, occlusion, and motion blur, together with evaluation code. Both are structurally close to DABACO---planar targets, video, four annotated corners---yet their targets are planar objects in general---a category that may well include screens---and their protocols evaluate planar object tracking: neither defines the active area of a display, and neither reports a pointing coordinate. Domain-specific homography benchmarks follow the same pattern: the SHL dataset~\cite{shl2025} annotates homographies for ice-hockey broadcast video, and the dynamic-scene homography dataset of Le et al.~\cite{le2020dynamic} couples homography estimation with per-pixel motion masks; neither involves a screen, an embedded platform, or a pointing task. Learned homography estimation~\cite{detone2016deep} addresses the geometric object directly---regressing the four corner displacements between two images---but is trained and evaluated on generic scene pairs rather than on displays.

A second family comes from document analysis: document-rectification benchmarks~\cite{docunet2018,doc3d2019,warpdoc2022,dir3002023,smartdoc2015,docreal2024} provide planar or quadrilateral annotations, but they target paper documents and OCR-oriented quality rather than display localization, temporal pointing, or embedded acquisition.

Third, resources that do involve displays divide into two groups that should not be conflated. On one hand, single-image recapture: the ROSE Lab Recaptured Images dataset~\cite{roselab_recaptured} and the rebroadcast collection of Farid et al.~\cite{farid2018rebroadcast} photograph real displays with many camera models, and CMA~\cite{cma2024} models the pixel-level distortion of screen-recaptured documents; none of them annotates a quadrilateral. On the other hand, screen presence in video: WildCam~\cite{wildcam2022} detects whether a screen is present in low-resolution egocentric video captured with an activity-oriented wearable camera, annotating presence and, in part, bounding boxes. Neither group provides the precise corners of the active area, and neither reports a pointing coordinate. Large pointing datasets also exist~\cite{deepoint2023}, but they annotate human pointing observed from a third-person view rather than the pointing coordinate of a device aimed at a display.

Table~\ref{tab:related-datasets} compares DABACO with these resources under checkable criteria. \emph{Video} marks resources with sequences of consecutive frames rather than independent images, and \emph{Sequence} marks ordered image sequences that are not necessarily temporally captured video; \emph{Temporal} denotes annotations provided per frame or with a frame index; \emph{Display} denotes that the annotated object is a display photographed by a camera, with \emph{Partial} meaning that displays may occur among the targets but are not the dataset's defined target category; \emph{Homography} is \textbf{Yes} when a projective map is distributed as such, \textbf{Derivable} when only the four corresponding corners are released (from which a homography can be computed), \emph{Partial} when the task estimates a more general deformation than a homography rather than distributing full homographies, and \textbf{No} otherwise; \emph{Corners} denotes annotation of the four corners of the active screen area (or of the document sheet for document benchmarks) rather than a box or an image-level class; \emph{Embedded} refers to acquisition on an embedded platform (acquisition only, not embedded \emph{evaluation}); \emph{Pointing} indicates that a pointing coordinate and its error are part of the released protocol; and \emph{Eval.\ code} indicates an official repository with standardized evaluation scripts.

\begin{table*}[t]
\centering
\footnotesize
\setlength{\tabcolsep}{6pt}
\caption{Comparison between related resources and DABACO. Column criteria are defined in the text.}
\label{tab:related-datasets}
\resizebox{\textwidth}{!}{%
\begin{tabular}{@{}lcccccccc@{}}
\toprule
\textbf{Dataset} & \textbf{Video} & \textbf{Temporal} & \textbf{Display} & \textbf{Homo-} & \textbf{Corners} & \textbf{Embedded} & \textbf{Pointing} & \textbf{Eval.} \\
 & & & & \textbf{graphy} & & & & \textbf{code} \\
\midrule
HPatches~\cite{balntas2017hpatches} & Sequence & Yes & No & Yes & No & No & No & Yes \\
HEB/Pi3D~\cite{barath2023large} & No & No & No & Yes & No & No & No & Yes \\
Screen Detection YOLO~\cite{su2023screen} & No & No & Yes & No & No & No & No & No \\
Screen Detection v2~\cite{screen-detection-v2_dataset} & No & No & Yes & No & No & No & No & No \\
PlanarTrack~\cite{planartrack2023,planartrack2025} & Yes & Yes & Partial & Derivable & Yes & No & No & Yes \\
POT-210~\cite{pot2018} & Yes & Yes & No & Derivable & Yes & No & No & Yes \\
SHL~\cite{shl2025} & Yes & Yes & No & Yes & No & No & No & Yes \\
Document benchmarks~\cite{docunet2018,doc3d2019,warpdoc2022,dir3002023,smartdoc2015,docreal2024} & No & No & No & Partial & Yes & No & No & Yes \\
Recapture~\cite{roselab_recaptured,farid2018rebroadcast,cma2024} & No & No & Yes & No & No & No & No & Yes \\
WildCam~\cite{wildcam2022} & Yes & Yes & Yes & No & No & Yes & No & Partial \\
DP Dataset / DeePoint~\cite{deepoint2023} & Yes & Yes & No & No & No & No & Yes & Yes \\
TUM rolling-shutter~\cite{tum_rolling_shutter} & Yes & Yes & No & No & No & Yes & No & Yes \\
\midrule
\textbf{\textit{DABACO}} & \textbf{Yes} & \textbf{Yes} & \textbf{Yes} & \textbf{Derivable} & \textbf{Yes} & \textbf{Yes} & \textbf{Yes} & \textbf{Yes} \\
\bottomrule
\end{tabular}}
\end{table*}

Table~\ref{tab:related-datasets} summarizes the comparison. To the best of our knowledge, no existing resource jointly provides physical-display video, exact active-area corners from which the homography is computed, temporal continuity, native capture on heterogeneous low-cost embedded hardware, pointing-oriented metrics, and a public toolkit with reference baselines. DABACO is designed to fill precisely that gap, providing a standardized foundation for developing embedded, camera-based pointing systems.

\section{The DABACO Dataset}
\label{sec:dataset}
The DABACO Dataset is a collection of annotated video sequences specifically designed for the development and evaluation of screen detection and pointing estimation algorithms under the constraints of low-cost embedded systems. It provides controlled real-world captures covering multiple display devices, visual contents, camera sensors, and acquisition settings. Crucially, the data were acquired with different camera sensors and edge-computing platforms, so the evaluation pipeline is exposed to the hardware and acquisition constraints relevant to low-cost interactive devices. Each sequence is provided with frame-level ground-truth annotations for the display's geometric corners, enabling reproducible evaluation of geometric and pointing performance.

\subsection{Hardware Setup and Capture Scenarios}
The dataset was captured using a Raspberry Pi 4 Model B and an ESP32-S3-WROOM CAM, combined with three low-cost camera modules, in order to expose the evaluation process to acquisition conditions relevant to low-cost edge devices. Algorithms of this kind must be validated against camera heterogeneity (e.g., monochrome global shutter versus color rolling shutter), varying resolutions and frame rates, and severe memory and CPU limitations, so data captured natively on such hardware is more representative than general video-tracking or VR pose datasets. Energy and latency on target hardware are deliberately outside the scope of this dataset, as they are properties of a specific implementation rather than of the recorded data; the measurement campaign that will report them is outlined in Section~\ref{sec:conclusion}.

Image acquisition and optical resolutions are governed by the specific camera modules:
\begin{itemize}
    \item \textbf{Arducam 1MP OV9281 (Monochrome Global Shutter):} Mounted on the Raspberry Pi 4. Operating up to high frame rates without rolling shutter distortion, crucial for fast pointing motions and accurate corner localization.
    \item \textbf{Arducam 5MP OV5647 (Color Rolling Shutter):} Mounted on the Raspberry Pi 4 with a standard M12 lens. Note that while the module supports interchangeable lenses, only the standard lens was used for this dataset.
    \item \textbf{ESP32-S3 Camera Module (OmniVision OV3660 3MP):} An embedded CMOS sensor module testing ultra-low-cost edge acquisition under constrained bandwidth and memory buffers.
\end{itemize}

\subsection{Ground-Truth Generation Pipeline}
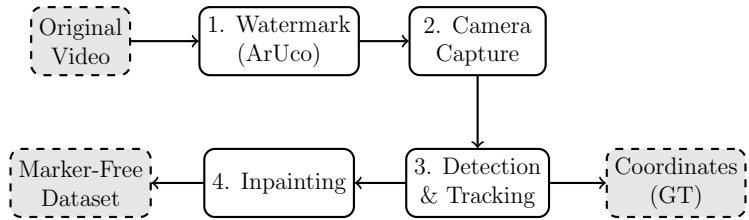
\begin{figure}[htpb]
    \centering
    \begin{tikzpicture}[
        node distance=2.8cm, 
        auto, 
        scale=0.75, 
        transform shape,
        box/.style={draw, rectangle, rounded corners, minimum height=1.2cm, minimum width=2.4cm, align=center, thick},
        io/.style={draw, rectangle, rounded corners, fill=gray!20, minimum height=1.2cm, align=center, thick, dashed}
        ]
        
        % Main blocks
        \node[io] (input) at (0,0) {Original\\Video};
        \node[box] (watermark) at (3.5,0) {1. Watermark\\(ArUco)};
        \node[box] (record) at (7.0,0) {2. Camera\\Capture};
        
        \node[box] (track) at (7.0,-2.5) {3. Detection\\\& Tracking};
        \node[box] (inpaint) at (3.5,-2.5) {4. Inpainting};
        
        % Outputs
        \node[io] (out_gt) at (10.5,-2.5) {Coordinates\\(GT)};
        \node[io] (out_video) at (0,-2.5) {Marker-Free\\Dataset};
        
        % Arrows
        \draw[->, thick] (input) -- (watermark);
        \draw[->, thick] (watermark) -- (record);
        \draw[->, thick] (record) -- (track);
        \draw[->, thick] (track) -- (inpaint);
        
        \draw[->, thick] (track) -- (out_gt);
        \draw[->, thick] (inpaint) -- (out_video);
        
    \end{tikzpicture}
    \caption{Overview of the Ground Truth Generation Pipeline.}
    \label{fig:pipeline}
\end{figure}

Generating rigorous ground truth for screen localization involves a trade-off. Synthetic datasets apply known projective transformations, but they struggle to faithfully simulate real physical captures affected by optical distortion, rolling shutter artifacts, motion blur, exposure dynamics, sensor noise, and display processing; annotating every frame of a physical capture manually, in turn, is labor-intensive and error-prone. We therefore designed a semi-automated pipeline that produces accurate screen-corner annotations across thousands of video frames while keeping human intervention to a minimum. The pipeline consists of four stages, illustrated in Figure~\ref{fig:pipeline} and summarized in Algorithm~\ref{alg:pipeline}.

\begin{figure*}[htpb]
    \centering
    \subfloat[Step 1: Watermarking]{\includegraphics[width=0.24\textwidth]{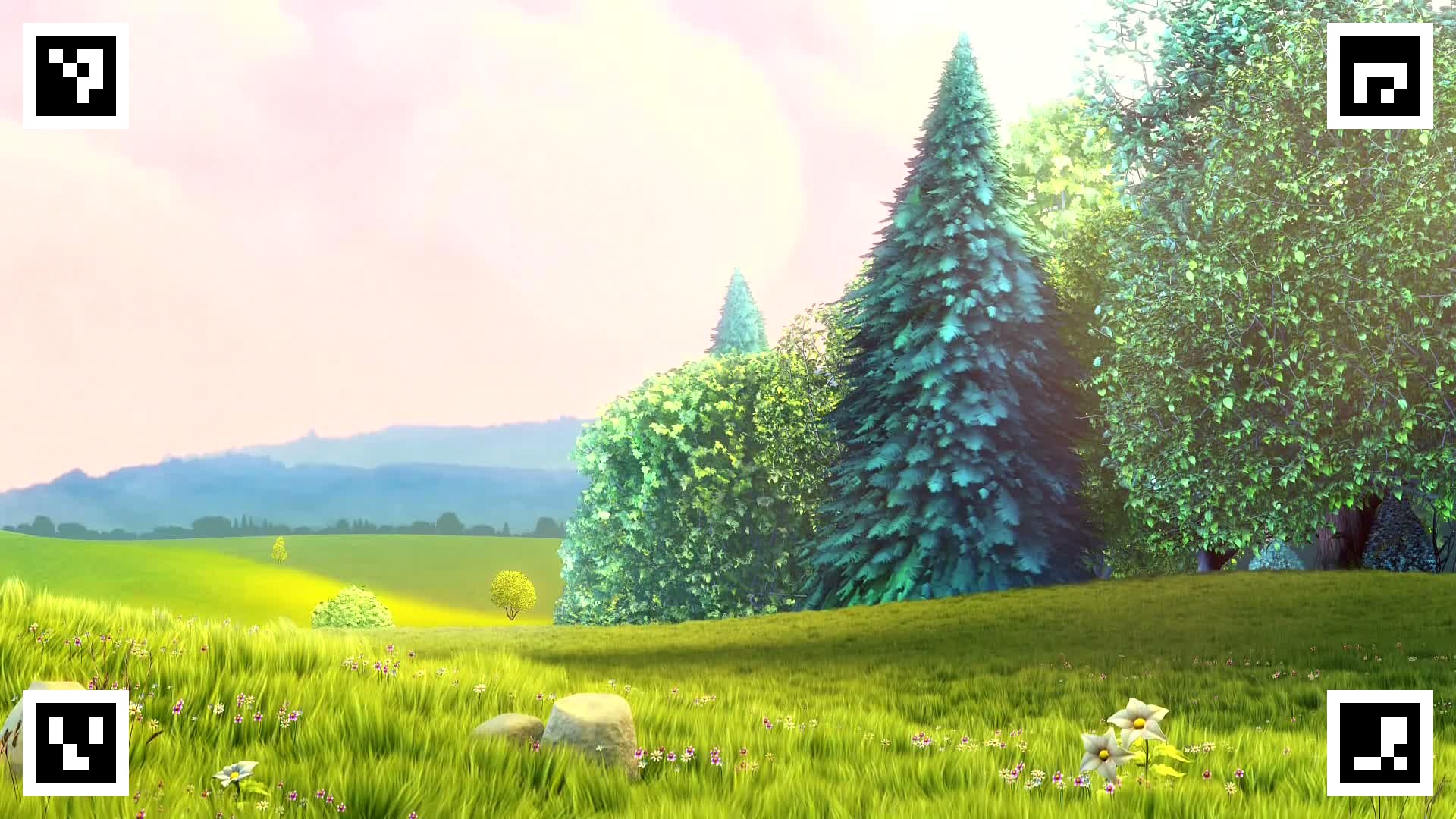}}\hfil
    \subfloat[Step 2: Camera Capture]{\includegraphics[width=0.24\textwidth]{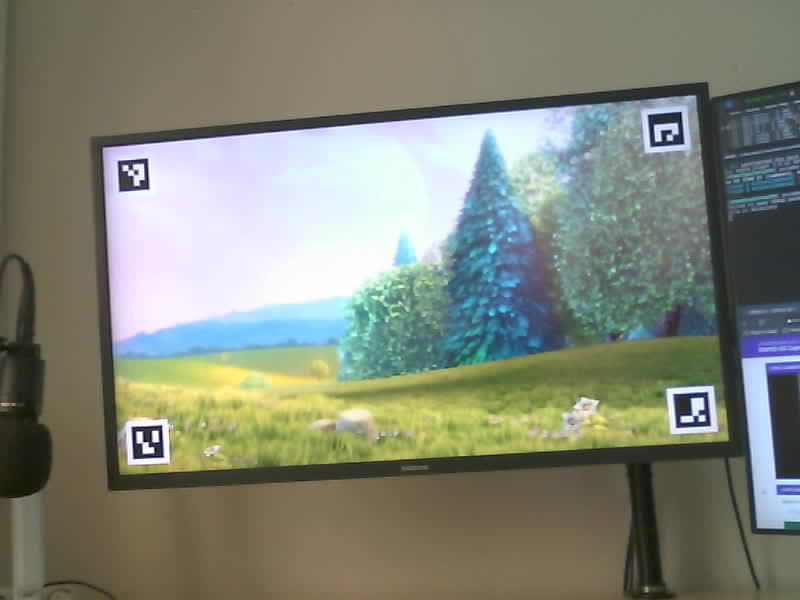}}\hfil
    \subfloat[Step 3: Detection \& Tracking]{\includegraphics[width=0.24\textwidth]{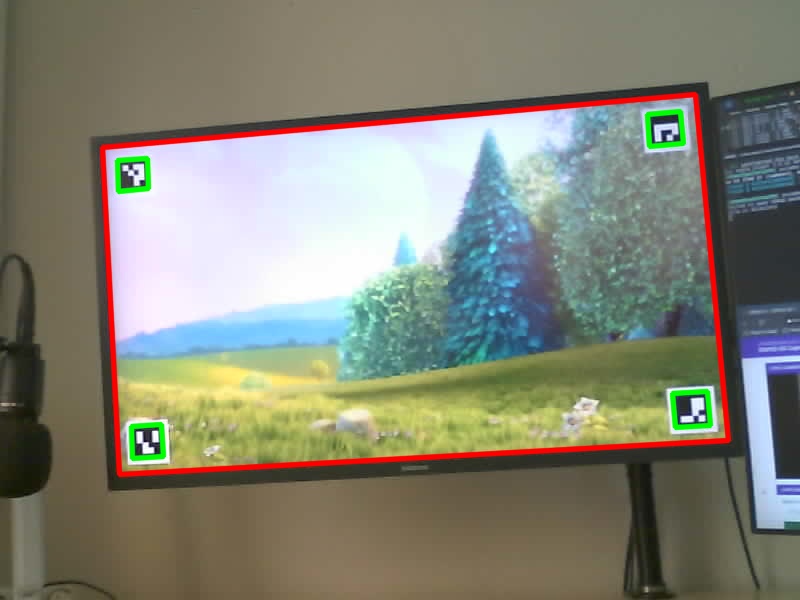}}\hfil
    \subfloat[Step 4: Telea Inpainting]{\includegraphics[width=0.24\textwidth]{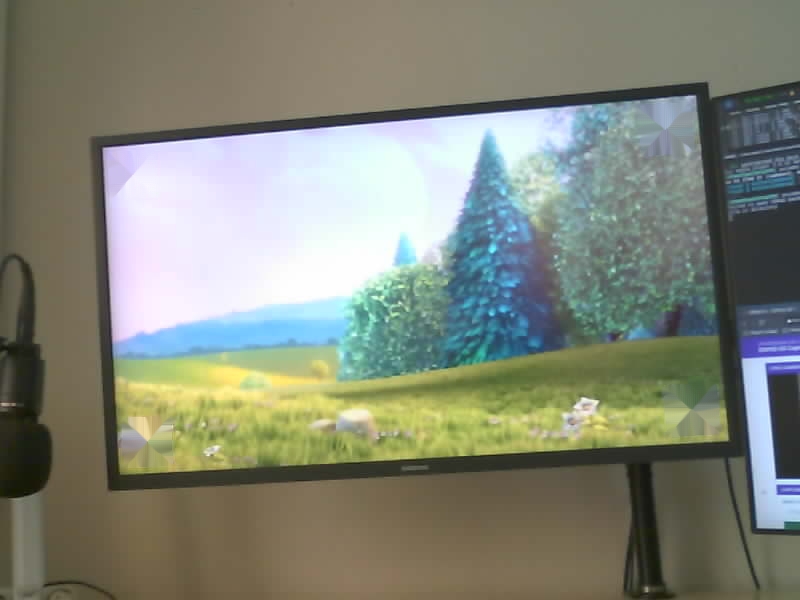}}
    \caption{Visual progression of a frame through the four stages of the Ground Truth Generation Pipeline.}
    \label{fig:pipeline_steps}
\end{figure*}

\begin{algorithm}[htpb]
\caption{Automatic Pre-annotation via Optical Flow Propagation}
\label{alg:pipeline}
\begin{algorithmic}[1]
\STATE \textbf{Input:} Video sequence $V = \{I_1, I_2, \dots, I_N\}$
\STATE \textbf{Output:} Corner coordinates $C = \{C_1, C_2, \dots, C_N\}$
\FOR{$t = 1$ \TO $N$}
    \STATE $C_t \leftarrow \text{DetectArUco}(I_t)$
    \IF{$\text{Confidence}(C_t) \ge \tau$}
        \STATE Mark $t$ as Anchor
    \ELSE
        \STATE Mark $t$ as Unknown
    \ENDIF
\ENDFOR
\FOR{each missing gap between consecutive Anchors $(t_A, t_B)$}
    \STATE $C_{A\rightarrow B} \leftarrow \text{ForwardLK}(V, C_{t_A}, t_A, t_B)$
    \STATE $C_{B\rightarrow A} \leftarrow \text{BackwardLK}(V, C_{t_B}, t_B, t_A)$
    \STATE $C_{t} \leftarrow \text{Blend}(C_{A\rightarrow B}, C_{B\rightarrow A}) \quad \forall t \in (t_A, t_B)$
\ENDFOR
\STATE \textbf{return} $C$
\end{algorithmic}
\end{algorithm}

\textbf{1. Source video watermarking.} Before the physical capture process, we synthetically embed four ArUco markers (\texttt{DICT\_4X4\_50})~\cite{garrido2014automatic} into the four inner corners of the active display area of the source video that will be played on the target display, so that each detected marker centre coincides with an active-area corner rather than with the bezel, frame, or margin of the device. Each marker includes an isolating white border (a quiet zone) to guarantee high contrast against dynamic video content, which ensures robust intra-frame detection by the camera.

\textbf{2. Detection and optical-flow propagation.} We execute an automatic pre-annotation step that combines a multi-pass ArUco detector with bidirectional Lucas--Kanade optical flow~\cite{lucas1981iterative}. As outlined in Algorithm~\ref{alg:pipeline}, the detector establishes high-confidence geometric anchors, filtering out collinear or unstable detections. For frames where markers are occluded, suffer from motion blur, or are distorted by rolling shutter effects, the bidirectional optical flow temporally tracks the missing corners from the nearest valid anchor frames. Any frame where one or more estimated corners fall outside the camera's field of view is explicitly marked as unlabeled.

\begin{figure*}[htpb]
    \centering
    \subfloat[Automatic Pre-annotation State]{\includegraphics[width=0.48\textwidth]{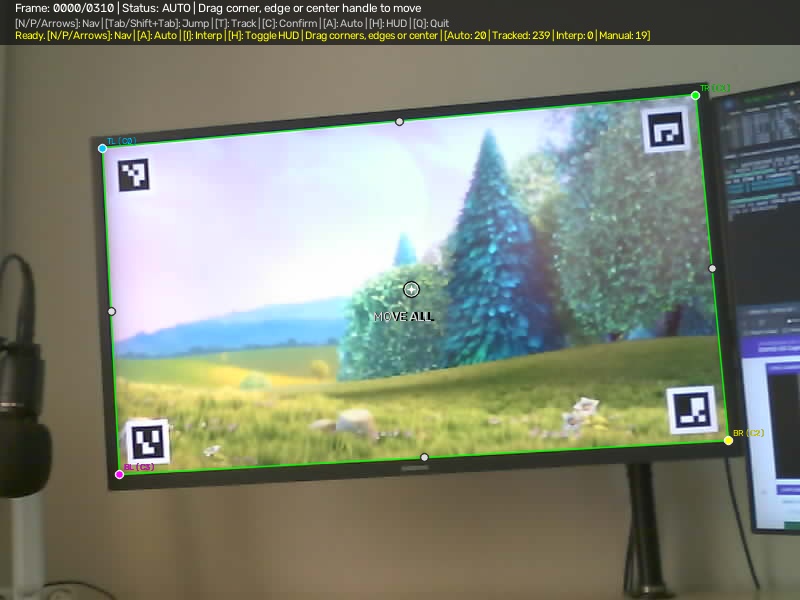}}\hfil
    \subfloat[Manual Correction State (with zoom loupe)]{\includegraphics[width=0.48\textwidth]{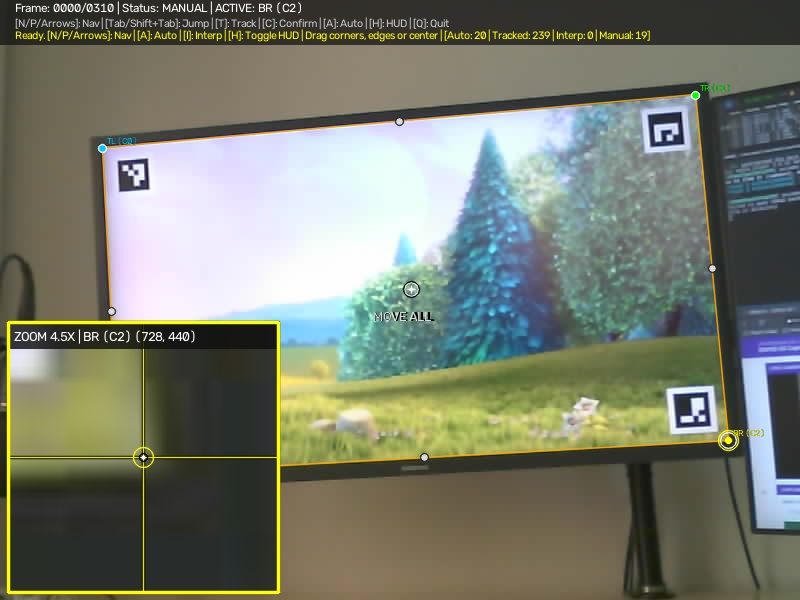}}
    \caption{The Interactive Verification editor's graphical user interface. The left panel shows the system's output during automatic pre-annotation. The right panel illustrates the manual review process, where annotators utilize the integrated $4.5\times$ zoom loupe to perform fine-grained coordinate corrections.}
    \label{fig:editor_ui}
\end{figure*}

\textbf{3. Interactive verification and manual correction.} To guarantee accuracy, a human annotator reviews the pre-annotated sequence using a custom interactive editor; the current labels were verified by a single annotator. Because manual verification is inherently error-prone, the tool provides a $4.5\times$ zoom loupe (Figure~\ref{fig:editor_ui}) for precise coordinate adjustments and smart navigation to jump directly to the frames that require review. When the annotator manually corrects a drifted corner, the system instantly re-propagates the optical flow trajectory using the new coordinate as an anchor.

\textbf{4. Marker removal via inpainting.} To prevent machine learning models from using the artificial markers as trivial detection cues rather than learning the actual display geometry, we digitally remove the markers and their quiet zones from the final images. We apply Telea inpainting~\cite{telea2004image} specifically localized using the ground-truth corner coordinates. Figure~\ref{fig:pipeline_steps} illustrates this full visual progression. To support dataset auditing, we release the original captures, the inpainted images, and the explicit modification masks.

\subsubsection{Pipeline Evaluation and Labeling Statistics}
The ground-truth generation toolkit is implemented entirely in Python, utilizing the OpenCV library~\cite{opencv_library} for all core computer vision routines. This ensures broad cross-platform compatibility and efficient matrix operations. The initial automatic pre-annotation phase is highly efficient: processing a standard $1920 \times 1080$ video sequence operates at approximately 15--20 frames per second on a standard desktop CPU. As a result, a typical 30-second capture (up to 1800 frames depending on the hardware framerate) is fully evaluated and pre-annotated in under two minutes without requiring hardware acceleration.

\begin{table}[t]
\centering
\caption{Frame annotation origins in the initial release.}
\label{tab:frame-counts}
\begin{tabular}{lrr}
\hline
\textbf{Annotation Source} & \textbf{Frames} & \textbf{Percentage} \\
\hline
Direct Marker Detection (Anchors) & 16,993 & 52.08\% \\
Optical Flow Propagation & 14,158 & 43.39\% \\
Manual Correction & 489 & 1.50\% \\
Discarded / Invalid Geometry & 989 & 3.03\% \\
\hline
\textbf{Total Processed Frames} & \textbf{32,629} & \textbf{100.00\%} \\
\hline
\end{tabular}
\end{table}

To evaluate the effectiveness of this pipeline, we aggregated the annotation statistics across the entire initial dataset release (27 sequences, yielding a total of 32,629 processed frames), as detailed in Table~\ref{tab:frame-counts}. Overall, 95.47\% of the processed frames received an automatic pre-annotation through either direct marker detection or bidirectional optical-flow propagation. This percentage describes the origin of the annotation, not an independently measured accuracy. 

The automatically generated annotations were subsequently reviewed using the interactive editor, with manual correction required for 489 frames (1.50\%). During this verification process, if a drift is detected due to severe reflections or extreme motion blur, the annotator manually corrects the affected frame, and the system instantly recalculates the Lucas--Kanade optical flow to re-propagate the correction. This substantially reduces the number of frames requiring direct manual editing compared with exhaustive frame-by-frame annotation.

\subsubsection{Inpainting Bias and Artifact Analysis}
While Telea inpainting successfully removes the stark visual contrast of the ArUco markers, it introduces localized artificial textures that could serve as unintended detection cues for high-capacity neural networks. In our pipeline, each source marker is rendered at $140 \times 140$ pixels on a standard $1920 \times 1080$ display, surrounded by a 12\% white quiet zone to guarantee contrast. During removal, the binary mask is expanded with a dilation of 2 pixels, and Telea inpainting is applied using a 1-pixel radius. 

\begin{figure}[htpb]
    \centering
    \subfloat[Original (Smooth)]{\includegraphics[width=0.28\textwidth]{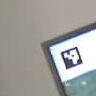}}\hfil
    \subfloat[Original (Structure)]{\includegraphics[width=0.28\textwidth]{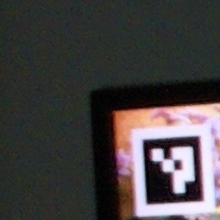}}\hfil
    \subfloat[Original (Border)]{\includegraphics[width=0.28\textwidth]{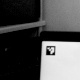}}\\
    \subfloat[Low-artifact interpolation]{\includegraphics[width=0.28\textwidth]{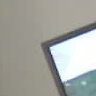}}\hfil
    \subfloat[Directional interpolation artifact]{\includegraphics[width=0.28\textwidth]{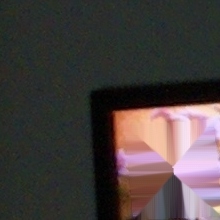}}\hfil
    \subfloat[Boundary bleeding artifact]{\includegraphics[width=0.28\textwidth]{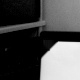}}
    \caption{Visual comparison of Telea inpainting across representative capture scenarios. Top row: original crops containing the ArUco markers. Bottom row: corresponding inpainted regions. Left (a, d): Seamless marker removal on uniform background (\texttt{esp32\_ov3660\_scrA\_v2\_m2vga}, frame 170). Middle (b, e): Stretched pixel structure forming a square of four converging triangles (\texttt{rpi4\_ov5647\_scrA\_v1\_m1fhd}, frame 94). Right (c, f): Inward bleeding of the dark screen bezel creating an artificial vignetting corner cut under reduced scale (\texttt{rpi4\_ov9281\_scrC\_v1\_m3hifps}, frame 267).}
    \label{fig:inpaint_artifacts}
\end{figure}

The visual fidelity of the inpainted regions exhibits distinct characteristics depending on background complexity and screen geometry. These observations are consistent with the general limitations of image inpainting methods, which may propagate nearby structures into the missing region~\cite{guillemot2014image}. The three artifact categories below were identified empirically in the DABACO captures (see Figure~\ref{fig:inpaint_artifacts}):
\begin{enumerate}
    \item \textbf{Uniform backgrounds (Low-artifact interpolation):} When markers fall on relatively smooth or uniform areas, the boundary pixels blend seamlessly, leaving virtually no perceptible trace (\textit{e.g.}, Figure~\ref{fig:inpaint_artifacts}(a, d)).
    \item \textbf{Directional interpolation (Geometric artifact):} On moderately textured backgrounds, pixel propagation from the four outer rectangular boundaries can form a subtle structure of stretched pixels that converge at the center, resembling a square partitioned into four triangular sectors (\textit{e.g.}, Figure~\ref{fig:inpaint_artifacts}(b, e)).
    \item \textbf{Boundary bleeding (Vignetting):} When the display is captured at a smaller scale or steep perspective where the marker lies in immediate proximity to the physical display bezel, the dark border pixels are pulled inward across the mask boundary, creating a pronounced dark diagonal corner cut or artificial vignetting effect (\textit{e.g.}, Figure~\ref{fig:inpaint_artifacts}(c, f)).
\end{enumerate}

To reduce the possibility that models exploit the artificial markers as shortcut cues rather than learning the display geometry, it is crucial to grant researchers full transparency and control over the training data. Consequently, alongside the marker-free, inpainted video sequences, the DABACO Dataset release explicitly provides the original video frames containing the physical ArUco markers, as well as the precise binary masks used during the inpainting step. This dual-release strategy empowers researchers to train and validate models using the inpainted images, or to apply alternative, mask-based data augmentation strategies (such as pure blackout masking or noise injection) to ensure models learn the actual geometric boundaries of the display rather than the artifacts left by the marker removal process.

\subsection{Dataset Structure and Statistics}
The dataset comprises 27 video sequences arranged in a controlled capture matrix: three cameras (an Arducam OV9281 monochrome global-shutter, an Arducam OV5647 color rolling-shutter, and an ESP32-S3 OV3660 IoT camera), three physical displays (a desktop monitor, a television, and a laptop screen), and three content sources: an animated film (Big Buck Bunny~\cite{blender2008bbb}), first-person voxel gameplay (MineClone 2~\cite{mineclone2}), and an academic presentation. The contents were chosen because their publicly documented licenses permit the intended use and because they cover variations in motion, color gradients, text density, and static overlays.

For each camera, we defined three operational modes that trade spatial detail against frame rate and motion blur: a \emph{high-quality} (HQ) mode, a \emph{medium-quality} (MQ) mode with shorter exposures, and a \emph{high-speed} (HS) mode that reduces resolution to maximize throughput. The specific resolution, frame rate, and exposure of each mode are camera-dependent and are listed in Table~\ref{tab:capture_matrix}. Because capturing every possible permutation of source videos and modes was unfeasible given the effort required for manual verification, each source was assigned exactly one mode per display, so that each camera--display pair is paired once with each content source under each of the three modes, without generating all mode permutations. This produces a balanced $3 \times 3 \times 3$ (camera $\times$ display $\times$ content) matrix of 27 sequences, with individual clips lasting approximately 30 seconds; the Latin Square assignment is detailed in Table~\ref{tab:assignment}.

\begin{table}[htpb]
\centering
\caption{Operational modes per camera: the three exposure and resolution configurations (HQ, MQ, HS).}
\label{tab:capture_matrix}
\resizebox{\columnwidth}{!}{%
\begin{tabular}{l l c c c}
\hline
\textbf{Camera} & \textbf{Type} & \textbf{High-Quality (HQ)} & \textbf{Medium-Quality (MQ)} & \textbf{High-Speed (HS)} \\
\hline
OV9281 & Mono, Global & $1280\times800$@30, Auto & $1280\times800$@30, 1/1000s & $640\times400$@100, Auto \\
OV5647 & Color, Rolling & 1080p@30, Auto & 960p@30, 1/500s & 720p@60, 1/1000s \\
ESP32-S3 & IoT, Color & $800\times600$@17.4, Auto & $640\times480$@24, 1/125s & $320\times240$@27.6, Auto \\
\hline
\end{tabular}%
}
\end{table}

\begin{table}[htpb]
\centering
\caption{Balanced Latin Square assignment of the three modes to each display--content combination.}
\label{tab:assignment}
\resizebox{\columnwidth}{!}{%
\begin{tabular}{l l c c c}
\hline
\textbf{Camera} & \textbf{Display} & \textbf{V1 (Animation)} & \textbf{V2 (Gameplay)} & \textbf{V3 (Presentation)} \\
\hline
OV9281 & Monitor & HQ & MQ & HS \\
 & TV & MQ & HS & HQ \\
 & Laptop & HS & HQ & MQ \\
\hline
OV5647 & Monitor & HQ & MQ & HS \\
 & TV & MQ & HS & HQ \\
 & Laptop & HS & HQ & MQ \\
\hline
ESP32-S3 & Monitor & HQ & MQ & HS \\
 & TV & MQ & HS & HQ \\
 & Laptop & HS & HQ & MQ \\
\hline
\end{tabular}%
}
\end{table}

This manageable scale is intentional for a preliminary dataset for low-resource devices: it allows us to prototype the annotation pipeline, establish baseline metrics, and identify algorithmic edge cases, and it can be extended with longer captures and more challenging scenarios in future iterations. Annotations are provided in JSON format, containing the exact pixel coordinates of the four screen corners for each frame. The release distributes these corner annotations only: no homography matrix is stored in the annotation files. The toolkit computes the homography from the four corners at evaluation time, in the canonical order top-left, top-right, bottom-right, bottom-left, which matches the order of the ground-truth annotations. This structured format facilitates direct parsing for evaluating intersection over union (IoU) and homography-based pointing estimation.

Sequence names follow the pattern \texttt{camera\_display\_video\_mode}; the display field takes the values \texttt{scrA} (desktop monitor), \texttt{scrB} (television), and \texttt{scrC} (laptop screen). In the mode field, the first token is the mode index used in Table~\ref{tab:capture_matrix}---\texttt{m1} for HQ, \texttt{m2} for MQ, and \texttt{m3} for HS---followed by a short code for the configuration actually captured: \texttt{std} or \texttt{fhd} (auto exposure), \texttt{fast} (short exposure), \texttt{hifps} (high frame rate), or \texttt{svga}, \texttt{vga}, \texttt{qvga} for the three ESP32-S3 modes. For example, \texttt{rpi4\_ov5647\_scrA\_v1\_m1fhd} is the OV5647, monitor, animation, high-quality 1080p capture, and \texttt{esp32\_ov3660\_scrB\_v2\_m3qvga} is the ESP32-S3, television, gameplay, high-speed QVGA capture. Each sequence directory contains the frame-level JSON annotation together with the capture metadata reported by the acquisition platform (resolution, frame rate, exposure, format, and frame count), so that the acquisition conditions of every clip are auditable from the release itself.

\section{Baselines for Screen Localization and Pointing}
\label{sec:baselines}
We provide two baseline algorithms to establish initial benchmarks on the DABACO dataset, available in our open-source toolkit.

\subsection{Classical Screen-Detection Baseline}
\begin{figure}[htpb]
    \centering
    \resizebox{\columnwidth}{!}{%
    \begin{tikzpicture}[
        box/.style={draw, rectangle, rounded corners, minimum width=2.5cm, minimum height=1cm, text centered, font=\small, fill=blue!10},
        arrow/.style={thick, ->, >=stealth}
    ]
        \node (input) at (0, 0) [box, fill=gray!10] {Input Frame};
        \node (filter) at (3.5, 0) [box] {Bilateral Filter};
        \node (edges) at (7.5, 0) [box, text width=2.5cm] {Canny Edges \&\\Thresholding};
        \node (contours) at (11.5, 0) [box, text width=2.5cm] {Contour\\Extraction};
        
        \node (geom) at (9.5, -2) [box, text width=2.5cm] {Convex Quad\\Filtering};
        \node (score) at (5.5, -2) [box, text width=2.5cm] {Aspect Ratio\\Scoring};
        \node (output) at (1.5, -2) [box, fill=green!10] {Detected Corners};

        \draw [arrow] (input) -- (filter);
        \draw [arrow] (filter) -- (edges);
        \draw [arrow] (edges) -- (contours);
        \draw [arrow] (contours) |- (11.5, -1) -| (geom);
        \draw [arrow] (geom) -- (score);
        \draw [arrow] (score) -- (output);
    \end{tikzpicture}
    }
    \caption{Block diagram of the classical computer vision baseline for screen detection.}
    \label{fig:classical_pipeline}
\end{figure}
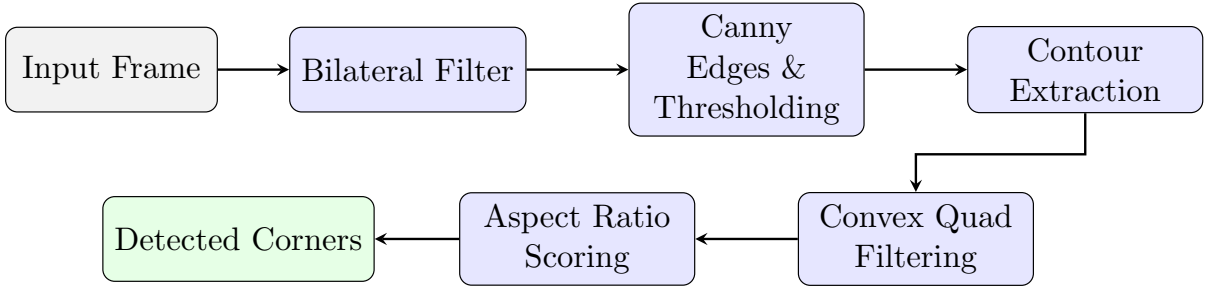

The classical baseline relies on geometric and intensity heuristics to extract the screen boundaries without requiring any learned parameters. The pipeline, illustrated in Figure~\ref{fig:classical_pipeline}, operates sequentially:

\begin{enumerate}
    \item \textbf{Preprocessing:} The input frame is converted to grayscale, and a bilateral filter is applied to reduce noise and internal screen textures while preserving sharp edge transitions.
    \item \textbf{Edge Extraction:} We compute multi-scale Canny edge maps combined with local adaptive thresholding to robustly extract contour maps across varying lighting conditions and screen brightness levels.
    \item \textbf{Contour Approximation:} Morphological closing is applied to bridge disconnected edges. The resulting contours are extracted and simplified into polygons using the Douglas-Peucker algorithm.
    \item \textbf{Geometric Filtering:} Candidate polygons are strictly filtered to retain only convex quadrilaterals (four vertices). 
    \item \textbf{Aspect Ratio Scoring:} The remaining candidates are scored based on their area, convexity, and structural similarity to standard screen aspect ratios (e.g., 16:9 or 16:10). The highest-scoring candidate is selected as the detected screen.
\end{enumerate}

\subsubsection*{Failure Cases and Limitations}
The classical pipeline excels in controlled environments with high contrast between the active screen and the background (e.g., a bright screen in a dimly lit room). However, it relies heavily on clear, continuous boundaries, making it prone to failure in several scenarios:
\begin{itemize}
    \item \textbf{Background Clutter:} Highly textured environments or rectangular objects (e.g., whiteboards, windows, or picture frames) can generate false positive quadrilaterals with higher geometric scores than the actual screen.
    \item \textbf{Partial Occlusions:} Objects partially blocking the screen boundary break the continuous Canny edges, preventing the Douglas-Peucker algorithm from forming a closed quadrilateral.
    \item \textbf{Reflections and Glare:} Strong reflections on the display surface disrupt the adaptive thresholding, causing the contour extraction to fragment the screen area.
    \item \textbf{Low-Contrast Borders:} If the screen displays dark content against a dark bezel and background, gradients become too weak to be reliably detected by the edge extraction step.
\end{itemize}

\subsection{Deep Learning-Based Object Detection Baseline}
As a modern counterpart to the classical approach, we provide a deep learning baseline utilizing a closed-vocabulary, off-the-shelf YOLOv8 instance segmentation model~\cite{Jocher_YOLO_by_Ultralytics_2023}.\footnote{Model: \texttt{yolov8s-seg.pt}, \url{https://docs.ultralytics.com/models/yolov8/}.} Notably, this model is deployed using its pre-trained COCO weights without any custom fine-tuning on the DABACO dataset. The model is restricted to the COCO \texttt{tv/monitor} class. Consequently, this baseline does not represent open-vocabulary screen segmentation and may fail on displays whose appearance differs from the training distribution, including laptop screens. It provides a reference measurement for the transfer of a closed-vocabulary segmentation model trained on COCO to the display-localization task without dataset-specific training.

The pipeline executes the following steps:
\begin{enumerate}
    \item \textbf{Preprocessing:} The input frame is resized to a standardized inference resolution (640 pixels along the longest edge) while preserving the original aspect ratio. The image is symmetrically padded with neutral gray to reach dimensions divisible by 32, which is required by the model's stride architecture.
    \item \textbf{Inference:} The YOLOv8 model generates bounding boxes and pixel-level segmentation masks. We filter detections to retain the \textit{tv} class at a very low confidence threshold (0.05) to prioritize recall over precision at the detection stage.
    \item \textbf{Candidate Selection:} Among all predicted instances, the detection encompassing the largest bounding box area is selected as the primary screen.
    \item \textbf{Polygon Approximation:} Direct extraction of the minimum area bounding box from the raw segmentation mask often introduces errors due to ``stairstepping'' artifacts along diagonal mask edges. To mitigate this, the boundary contour of the mask is iteratively simplified using the Douglas-Peucker algorithm, dynamically increasing the epsilon parameter (from 1\% to 10\% of the perimeter) until exactly four vertices are found. If a 4-point polygon cannot be extracted via simplification, the algorithm falls back to computing the minimum area bounding box of a simplified contour.
    \item \textbf{Coordinate Restoration:} Finally, the four estimated vertices are scaled back to the original image dimensions and sorted into a canonical order (top-left, top-right, bottom-right, bottom-left).
\end{enumerate}

While this deep learning baseline transfers a COCO-pretrained segmentation model without dataset-specific training and, in qualitative terms, tolerates partial occlusions, background clutter, and extreme lighting variations better than the hand-crafted classical heuristics, it operates at a higher computational cost and its advantage does not extend to the corner-level geometry required for pointing. Furthermore, because the predicted mask is derived from higher-level semantic features rather than sharp subpixel gradients, the resulting corner coordinates may exhibit structural misalignment and outliers that are particularly detrimental to homography-based pointing estimation.

\section{Evaluation Toolkit and Metrics}
\label{sec:evaluation}
Our toolkit provides a standardized evaluation pipeline to benchmark both the geometric accuracy of the detected screen and the functional accuracy of the derived pointing coordinates. It reports \emph{frame-level} metrics, which are computed per frame and aggregated over the evaluated frames, and one \emph{sequence-level} metric. Unless otherwise stated, aggregate metrics pool all evaluated frames across the 27 sequences so that each frame contributes equally, and frames discarded during ground-truth generation are excluded from every metric. A missed prediction is penalized by convention: it receives an IoU of $0$ and a normalized pointing error of $1.0$, so coverage and geometric quality jointly affect the aggregate values.

The frame-level metrics are:
\begin{itemize}
    \item \textbf{Detection Rate (Coverage, \%):} the proportion of frames in which the algorithm predicts a valid four-corner screen quadrilateral.
    \item \textbf{False-Lock Rate (\%):} the proportion of predictions whose normalized pointing error exceeds the threshold $\tau_p = 0.05$, computed only over frames in which the algorithm returns a prediction. Missed predictions are not counted as false locks; they reduce coverage instead.
    \item \textbf{Corner Error:} the mean Euclidean distance between the predicted and ground-truth corners, reported both in pixels and as a percentage of the image diagonal.
    \item \textbf{Intersection over Union (IoU):} the spatial overlap between the predicted polygon and the ground-truth quadrilateral.
    \item \textbf{Pointing Error (\%):} the homography-derived pointing-coordinate error. Let $\mathbf{H}$ be the homography that maps the screen plane to the image, computed from the corresponding quadrilateral corners. The image point at the centre of the captured frame,
    \[
    \mathbf{p}_{\mathrm{img}} = \Bigl(\tfrac{W}{2},\,\tfrac{H_{\mathrm{im}}}{2},\,1\Bigr)^{\mathsf T},
    \]
    where $W$ and $H_{\mathrm{im}}$ are the image width and height and no camera calibration is applied, is mapped to $\tilde{\mathbf{q}} = \mathbf{H}\,\mathbf{p}_{\mathrm{img}}$ and converted to normalized screen coordinates $\mathbf{q} = (\tilde q_x/\tilde q_z,\ \tilde q_y/\tilde q_z)$, with $(0,0)$ the top-left and $(1,1)$ the bottom-right corner. The error is the Euclidean distance between the coordinates obtained from the predicted and ground-truth quadrilaterals,
    \[
    e_{\mathrm{point}} = \bigl\lVert \mathbf{q}_{\mathrm{pred}} - \mathbf{q}_{\mathrm{GT}} \bigr\rVert_2,
    \]
    reported as a percentage by multiplying by 100. The value is a proxy for the pointing direction and should not be interpreted as the fully calibrated mechanical pointing error of the final device.
\end{itemize}

The sequence-level metric is:
\begin{itemize}
    \item \textbf{Time-to-Lock (frames):} the number of frames from the start of a sequence until the algorithm first returns a valid quadrilateral whose normalized pointing error is below $\tau_p = 0.05$, indicating the initial responsiveness of the system.
\end{itemize}

The fixed penalty of $1.0$ assigned to missed predictions is a convention, not an upper bound: a valid prediction whose homography is close to singular can project the image centre far outside the normalized screen space, so pointing error can exceed 100\% for geometrically unstable predictions. To characterize the error distribution rather than only its mean, the toolkit also reports the median and 95th-percentile pointing error, which should be read together with coverage.

\section{Results and Benchmarking}
\label{sec:results}
To establish a point of reference for future algorithms, we evaluated both the classical and the deep learning baselines across all 27 sequences in the DABACO dataset. Table~\ref{tab:results} reports coverage, false-lock rate, corner error, IoU, and pointing error aggregated over all evaluated frames, together with pointing error over valid predictions only and the mean time-to-lock. The toolkit additionally reports the median and 95th percentile for both aggregation protocols. The detailed frame-by-frame metrics and visualizations, including Virtual HUD projections, are accessible via the interactive HTML benchmark dashboard generated by the toolkit.

\begin{table*}[t]
    \centering
    \small
    \setlength{\tabcolsep}{3pt}
    \caption{Aggregate Baseline Performance on the DABACO Dataset (Mean values, with medians in parentheses). The ``all frames'' column reports the mean over all evaluated frames, including the default penalty for missing predictions. The ``valid preds.'' column reports the mean restricted to frames for which a valid prediction was returned. Median and P95 values for both aggregation protocols are available in the accompanying benchmark report.}
    \begin{tabularx}{\textwidth}{@{}l *{7}{>{\centering\arraybackslash}X}@{}}
        \toprule
        \textbf{Algorithm} & \textbf{Coverage (\%)} & \textbf{False-Lock (\%)} & \textbf{Corner Error (\%)} & \textbf{IoU (\%)} & \textbf{Pointing Error (\% -- all)} & \textbf{Pointing Error (\% -- valid)} & \textbf{Time-to-Lock (frames)} \\
        \midrule
        Classical & \textbf{92.63} & 58.79 & 12.52 (11.44) & 52.98 (53.24) & \textbf{28.74 (23.30)} & \textbf{23.07} & 39.7 \\
        YOLOv8 Seg. & 86.97 & \textbf{52.31} & \textbf{7.89 (6.39)} & \textbf{58.84 (63.16)} & 90.56 (30.72) & 89.15 & \textbf{25.4} \\
        \bottomrule
    \end{tabularx}
    \label{tab:results}
\end{table*}

\begin{table}[htpb]
\centering
\small
\setlength{\tabcolsep}{4pt}
\caption{Baseline performance by display type (Mean values reported). Pointing error can exceed 100\% for valid but geometrically unstable homography predictions.}
\label{tab:results-display}
\resizebox{\columnwidth}{!}{%
\begin{tabular}{llcccc}
\toprule
\textbf{Display} & \textbf{Method} & \textbf{Coverage (\%)} & \textbf{False-Lock (\%)} & \textbf{IoU (\%)} & \textbf{Pointing Error (\% -- all frames)} \\
\midrule
Monitor & Classical & 96.71 & 62.09 & 58.76 & 21.09 \\
Monitor & YOLOv8 & \textbf{97.13} & \textbf{35.35} & \textbf{76.99} & \textbf{9.83} \\
\addlinespace
TV & Classical & \textbf{89.86} & \textbf{37.08} & 67.56 & \textbf{19.23} \\
TV & YOLOv8 & 84.32 & 38.33 & \textbf{68.59} & 20.10 \\
\addlinespace
Laptop & Classical & \textbf{91.31} & \textbf{77.19} & \textbf{32.64} & \textbf{45.91} \\
Laptop & YOLOv8 & 79.47 & 83.27 & 30.93 & 241.76 \\
\bottomrule
\end{tabular}}
\end{table}

The classical baseline achieves a higher overall coverage (92.63\%) than the closed-vocabulary YOLOv8 baseline (86.97\%). This performance drop in the deep learning baseline is heavily influenced by the laptop scenarios in the dataset. Because our YOLOv8 pipeline was strictly configured to filter for the \textit{tv/monitor} class (COCO class 62), it struggles with laptop screens, which degrades its aggregate coverage. We deliberately retained only the COCO \texttt{tv/monitor} class to preserve the off-the-shelf, closed-vocabulary setting: introducing a dedicated laptop class would require a dataset-specific label mapping and would no longer represent the intended zero-training transfer baseline. This is therefore a limitation of the baseline rather than an intrinsic limitation of segmentation models. The model also occasionally fails to recognize screens if the displayed content deviates drastically from the typical appearance of the training data in MS COCO~\cite{lin2014microsoft}. When a display is detected, YOLOv8 achieves lower mean corner error (7.89\%) and higher mean IoU (58.84\%) than the classical baseline in this evaluation. However, this pixel-level advantage does not translate into accurate homography-derived pointing coordinates.

Indeed, the classical approach achieves a substantially lower aggregate pointing error (28.74\% mean, 23.30\% median) compared to YOLOv8 (90.56\% mean, 30.72\% median). The aggregate pointing error includes the penalty assigned to frames without a valid prediction. Therefore, the substantially higher YOLOv8 mean pointing error reflects both its geometric errors on valid detections and its lower coverage. The median provides a complementary view of the error distribution among the evaluated frames, while coverage must be considered jointly when interpreting this metric. This discrepancy arises from the extreme sensitivity of the homography matrix to structural misalignments. While the YOLO segmentation mask covers the screen area well, its derived corner coordinates often suffer from ``stairstepping'' artifacts and slight vertex drifts along the bezel. These small geometric inconsistencies severely skew the resulting perspective projection, generating extreme outliers that destroy the mean pointing accuracy. Conversely, the classical pipeline explicitly constrains the prediction to a convex quadrilateral, which can produce geometrically consistent corner configurations even when the estimated screen boundary has a relatively low pixel overlap with the annotation.

Table~\ref{tab:results-display} breaks the results down by display type and shows that the two baselines behave differently across screens. On monitors, YOLOv8 achieves better coverage, IoU, and pointing error than the classical pipeline; on televisions, the two methods are closer, with the classical baseline slightly ahead in coverage and pointing error; and on laptops, YOLOv8 degrades markedly because of the \texttt{tv/monitor} class restriction. Overall, the classical baseline is the more stable of the two across displays, but its advantage comes from producing geometrically consistent quadrilaterals rather than from a better pixel-level localization of the screen. More importantly, the pointing errors of both baselines remain substantial: even the better-performing method is far from the accuracy required for a practical pointing device. The task is therefore far from solved, especially under the low-resource and heterogeneous conditions targeted by DABACO.

\section{Conclusion and Future Work}
\label{sec:conclusion}
In this paper, we introduced the DABACO Dataset, a multi-camera video dataset designed specifically for evaluating screen detection and pointing estimation algorithms on embedded systems. By employing a semi-automated pipeline based on ArUco markers, optical-flow tracking, manual verification, and marker removal, we detail a systematic procedure for generating and validating geometric annotations across heterogeneous acquisition platforms (Raspberry Pi 4B, ESP32-S3). Alongside the dataset, we released an open-source evaluation toolkit and provided initial benchmarks comparing a classical computer vision pipeline and a closed-vocabulary, off-the-shelf deep learning model (YOLOv8). The results highlight a critical trade-off: while segmentation models achieve superior pixel-level overlap (IoU), classical geometric approaches currently provide more stable and reliable perspective projections for pointing estimation.

Future work will focus on four directions. First, we plan to expand the dataset with a wider variety of dynamic lighting conditions, extreme perspective distortions, and additional display types (such as smartphones and projectors), following an iterative process in which the cases that prove most challenging for the baselines guide the next captures.

Second, we aim to leverage this dataset to train and optimize lightweight, end-to-end neural architectures specifically tailored for ultra-low-power microcontrollers like the ESP32. By addressing the geometric instability of current segmentation masks and minimizing computational overhead, we hope to enable robust, real-time pointing estimation natively on edge computing devices.

Third, we will run a dedicated measurement campaign on target hardware, with its protocol fixed in advance: for each platform (Raspberry Pi 4B, Raspberry Pi Zero 2W, and ESP32-S3) a device-specific build will be evaluated under two conditions---\textbf{live capture}, which includes acquisition and transport in the power budget, and \textbf{deterministic replay from storage}, which feeds exactly the frames annotated in this release and therefore excludes the camera from that budget---reporting accuracy, latency (p50/p95), peak memory, and energy per frame and per pointing report. We will adapt the reporting discipline of MLPerf Tiny~\cite{mlperf2021tiny} for accuracy, latency, and energy, but the standardized MLPerf Tiny tasks do not include corner estimation or pointing, so DABACO results will not be presented as directly comparable with an MLPerf Tiny score. The reference classical pipeline also relies on OpenCV routines that are not available on the microcontroller and must be reimplemented, so the port itself is a substantial part of that work. This follows from the scoping decision stated in Section~\ref{sec:dataset}: energy and latency on target hardware are deliberately outside the scope of the current release.

Fourth, we will refine the annotation pipeline by decoupling the raw ArUco detection error from the optical-flow tracking drift and by formalizing metrics for the geometric accuracy of the labels and the visual quality of the inpainted regions, which will also enable the use of learning-based marker removal.

\section*{Acknowledgements}
This work has been funded by the \textit{Vicerrectorado de Investigación y Transferencia} of the \textit{Universidad de La Laguna} (ULL) under the 2025 research incentive program for early-career researchers (\textit{proyectos para el personal investigador novel}), project ``DABACO'', accounting file No. 2026/0000637 led by Óscar Gómez, and has also received support from the Canary Islands Agency for Research, Innovation, and Information Society (ACIISI) of the Government of the Canary Islands.

\section*{Generative AI Disclosure}
During the preparation of this work the authors used generative AI to aid with the redaction of this manuscript. After using this tool/service, the authors reviewed and edited the content as needed and take full responsibility for the content of the published article.

\bibliographystyle{unsrt}
\bibliography{references}

\end{document}